\documentclass{article}
\usepackage{ijcai20}

\usepackage{times}
\usepackage{soul}
\usepackage{url}
\usepackage[hidelinks]{hyperref}
\usepackage[utf8]{inputenc}
\usepackage[small]{caption}
\usepackage{graphicx}
\usepackage{amsmath}
\usepackage{amssymb}
\usepackage{amsthm}
\usepackage{booktabs}
\usepackage{algorithm}
\usepackage{algorithmic}
\usepackage{multirow}
\usepackage{color}
\usepackage{array}

\usepackage{pgfplots}
\usepackage{tikz}
\usepackage{subfigure}
\usepackage{latexsym}
\usepackage{tikz-qtree}
\usepackage{hyperref}
\usepgflibrary{arrows.meta}
\usetikzlibrary{fit, calc, decorations.pathreplacing, positioning, shapes.geometric}

\title{Fast and Accurate Neural CRF Constituency Parsing}

\author{
Yu Zhang\thanks{Yu Zhang and Houquan Zhou make equal contributions to this work. Zhenghua Li is the corresponding author.},
Houquan Zhou$^{\ast}$,
Zhenghua Li
\affiliations
Institute of Artificial Intelligence, School of Computer Science and Technology, \\ Soochow University, Suzhou, China \\
\emails
yzhang.cs@outlook.com,
hqzhou@stu.suda.edu.cn,
zhli13@suda.edu.cn
}
\begin{document}
\maketitle

\begin{abstract}
\label{section:abstract}

Estimating probability distribution is one of the core issues in the NLP field. However, in both deep learning (DL) and pre-DL eras, unlike the vast applications of linear-chain CRF in sequence labeling tasks, very few works have applied tree-structure CRF to constituency parsing, mainly due to the complexity and inefficiency of the inside-outside algorithm.
This work presents a fast and accurate neural CRF constituency parser. The key idea is to batchify the inside algorithm for loss computation by direct large tensor operations on GPU, and meanwhile avoid the outside algorithm for gradient computation via efficient back-propagation.
We also propose a simple two-stage bracketing-then-labeling parsing approach to improve efficiency further.
To improve the parsing performance, inspired by recent progress in dependency parsing, we introduce a new scoring architecture
based on boundary representation and biaffine attention, and a beneficial dropout strategy.
Experiments on PTB, CTB5.1, and CTB7 show that our two-stage CRF parser achieves new state-of-the-art performance on both settings of w/o and w/ BERT, and can parse over 1,000 sentences per second.
We release our code at https://github.com/yzhangcs/crfpar.

\end{abstract}
\section{Introduction}
\label{section:introduction}

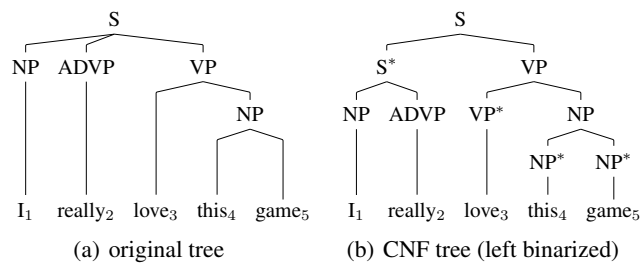
\begin{figure}[tb]
\subfigure[original tree]
{
\label{fig:const-tree-original}
  \begin{minipage}[b]{0.22\textwidth}
      \centering
      \begin{tikzpicture}[
          scale=0.8,
          level distance=22.5pt,
          every tree node/.style={align=center,anchor=base},
          frontier/.style={distance from root=90pt},
          edge from parent/.style={draw,edge from parent path={(\tikzparentnode.south) {[rounded corners=0.5pt]-- ($(\tikzchildnode |- \tikzparentnode.south) + (0, -5pt)$) -- (\tikzchildnode)}}}
          ]
      \Tree
      [.S
          [.NP I$_1$ ]
          [.ADVP really$_2$ ]
          [.VP love$_3$ [.NP this$_4$ game$_5$ ] ] ];
  \end{tikzpicture}

  \end{minipage}
}
\hfill
\subfigure[CNF tree (left binarized)]
{
\label{fig:binaried-tree}
  \begin{minipage}[b]{0.22\textwidth}
      \centering
      \begin{tikzpicture}[
          scale=0.8,
          level distance=22.5pt,
          every tree node/.style={align=center,anchor=base},
          frontier/.style={distance from root=90pt},
          edge from parent/.style={draw,edge from parent path={(\tikzparentnode.south) {[rounded corners=0.5pt]-- ($(\tikzchildnode |- \tikzparentnode.south) + (0, -5pt)$) -- (\tikzchildnode)}}}
        ]
      \Tree
          [.S
          [.$\textrm{S}^\ast$ [.NP I$_1$ ] [.ADVP really$_2$ ] ]
          [.VP
            [.$\textrm{VP}^\ast$ love$_3$ ]
            [.NP [.$\textrm{NP}^\ast$ this$_4$ ]
                 [.$\textrm{NP}^\ast$ game$_5$ ] ] ] ];
      \end{tikzpicture}
  \end{minipage}
}
\caption{
    Example constituency trees. Part-of-speech (POS) tags are not used  as inputs in this work and thus excluded.
    %due to simplicity and no hurt on performance.
    % (in Figure~\ref{fig:constituency-tree}) and its corresponding left binarized version (in Figure~\ref{fig:binaried-tree}).
}
\label{fig:const-tree-full-figure}
\end{figure}

Given an input sentence, constituency parsing aims to build a hierarchical tree as depicted in Figure~\ref{fig:const-tree-original}, where the leaf or terminal nodes correspond to input words and non-terminal nodes are constituents (e.g., $\texttt{VP}_{3,5}$).
%In, constituency parsing
As a fundamental yet challenging task in the natural language processing (NLP) field, constituency parsing has attracted a lot of research attention since large-scale treebanks were annotated,  such as Penn Treebank (PTB), Penn Chinese Treebank (CTB), etc.
Parsing outputs are also proven to be extensively useful for a wide range of downstream applications \cite{akoury-etal-2019-syntactically,wang-etal-2018-tree}.
% 暂时不加参考文献了，太多了。
% like sematic parsing \cite{jiang-etal-2019-hlt,li-etal-2017-modeling,}, relation extraction \cite{qian-etal-2008-exploiting}, and machine translation .

As one of the most influential works, \citeauthor{collins-1997-three}~\shortcite{collins-1997-three} extends methods from probabilistic context-free grammars (PCFGs) to lexicalized grammars.
Since then, constituency parsing has been dominated by such generative models for a long time, among which
the widely used Berkeley parser adopts an unlexicalized PCFG with latent non-terminal splitting annotations \cite{matsuzaki-etal-2005-probabilistic,petrov-klein-2007-improved}.
As for discriminative models, there exist two representative lines of research.
The first line adopts the graph-based view based on dynamic programming decoding, using either local max-entropy estimation \cite{kaplan-etal-2004-speed} or global max-margin training \cite{taskar-etal-2004-max}.
The second group builds a tree via a sequence of shift-reduce actions based on greedy or beam decoding, known as the transition-based view \cite{sagae-lavie-2005-classifier,zhu-etal-2013-fast}.

% based on dynamic programming decoding as an alternative to maximum entropy estimation for  graph-based parsing.
% Henderson 2004, zhang yue 2009-IWPT

% BERKELY parsing speed
% Zpar speed
% Finkel-CRF is much slower than both.

%TODO
% 建Demo，短语依存MTL，或者只是短语。中英文都要有。分词和tokenization第一版不加入，让用户输入。后面考虑加入进来。
% 问题：现在英语的tokenization是输入给定的吗？

Recently, constituency parsing has achieved significant progress thanks to the impressive capability of deep neural networks in context representation.
Two typical and popular works are respectively the transition-based parser of \citeauthor{cross-huang-2016-span}~\shortcite{cross-huang-2016-span} and the graph-based parser of \citeauthor{stern-etal-2017-minimal}~\shortcite{stern-etal-2017-minimal}.
As discriminative models, the two parsers share several commonalities, both using 1) multi-layer BiLSTM as encoder; 2) minus features from BiLSTM outputs as span representations; 3) MLP for span scoring; 4) max-margin training loss.
Most works \cite{gaddy-etal-2018-whats,kitaev-klein-2018-constituency} mainly follow the two parsers and achieve much higher parsing accuracy than traditional non-neural models, especially with contextualized word representations trained with language modeling loss on large-scale unlabeled data \cite{peters-etal-2018-deep,devlin-etal-2019-bert}.

Despite the rapid progress, existing constituency parsing research suffers from two closely related drawbacks. First, parsing (also for training) speed is slow and can hardly satisfy the requirement of %for online parsing in
real-life systems.
Second, the lack of explicitly modeling tree/subtree probabilities may hinder the effectiveness of utilizing parsing outputs.
On the one hand, estimating probability distribution is one of the core issues in the NLP field \cite{le-zuidema-2014-inside}. On the other hand, compared with unbounded tree scores, tree probabilities can better serve high-level tasks as soft features \cite{jin-etal-2020-relation},
%are intuitively more useful than unbounded tree scores
and marginal probabilities of subtrees can support
the more sophisticated Minimum Bayes Risk (MBR) decoding \cite{smith-smith-2007-probabilistic}.

In fact, \citeauthor{finkel-etal-2008-efficient}~\shortcite{finkel-etal-2008-efficient} and \citeauthor{durrett-klein-2015-neural}~\shortcite{durrett-klein-2015-neural} both propose CRF-based constituency parsing by directly modeling the conditional probability. % $p(\boldsymbol{t}\mid\boldsymbol{x})$.
However, both models are extremely inefficient due to
the high time-complexity of the inside-outside algorithm for loss and gradient computation, especially the outside procedure.
The issue becomes more severe in the DL era since all previous works perform the inside-outside computation on CPUs according to our knowledge and switching between GPU and CPU is expensive.
%unmatched speed of CPU/GPU computations.

%lafferty也得引用！ 2001

This work proposes a fast and accurate CRF constituency parser by substantially extending the graph-based parser of \citeauthor{stern-etal-2017-minimal}~\shortcite{stern-etal-2017-minimal}. The key contribution is that we batchify the inside algorithm for direct loss and gradient computation on GPU. Meanwhile, we find that the outside algorithm can be efficiently fulfilled by automatic back-propagation, which is shown to be equally efficient with the inside (forward) procedure, naturally verifying the great theoretical work of \citeauthor{eisner-2016-inside}~\shortcite{eisner-2016-inside}.
Similarly, we batchify the Cocke–Kasami–Younger (CKY) algorithm for fast decoding.

In summary, we make the following contributions.
% \footnote{Our code is available at \url{https://github.com/yzhangcs/crfpar}}
\begin{itemize}
\item We for the first time propose a fast and accurate CRF constituency parser for directly modeling (marginal) probabilities of trees and subtrees.
The efficiency issue, which bothers the community for a long time, is well solved by elegantly batchifying the inside and CKY algorithms for direct computation on GPU.
% CRF batchify fully utilize the GPU power, directly compute on GPU. We for the first time introduce CRF to deep neural networks for constituency parsing, outperforming the max-margin methods.
% \footnote{To our knowledge, this is the first BiLSTM-based CRF parser for constituency parsing.}

% We propose to batchify the inside method, which makes it possible for parallelized large tensor computation on GPU. We also show that the sophisticated outside algorithm is no longer necessary, and can be replaced by the equally efficient back-propagation process for the computation of gradients and marginal probabilities.

\item We propose a two-stage bracketing-then-labeling parsing approach that is more efficient and achieves slightly better performance than the one-stage method.

%We borrow the architecture from biaffine parser, and exploit their advantages in the constituency parsing. All of these contribute to the fast (over 1,000 sentences per second) and accurate parsing.

\item We propose a new span scoring architecture based on span boundary representation and biaffine attention scoring, which performs better than the widely used minus-feature method. %scoring  better scoring architecture (biaffine) and better parameter settings such as dropouts. substantially improve performance.
We also show that the parsing performance can be improved by a large margin via better parameter settings such as dropout configuration.

\item Experiments on three English and Chinese benchmark datasets show that our proposed two-stage CRF parser achieves new state-of-the-art parsing performance under both settings of w/o and w/ BERT \cite{devlin-etal-2019-bert}).
In terms of parsing speed, our parser can parse over 1,000 sentences per second.

%     %In the closed scenario that does not use any external resources, all of them achieve the state-of-the-art parsing results. Release code and models at github.
%     Speed: 1,000 sentences per second.
% new state-of-the-art results under several settings on three benchmark datasets of English and Chinese.

\end{itemize}

\begin{figure}[tb]
\centering
\includegraphics [scale=0.8] {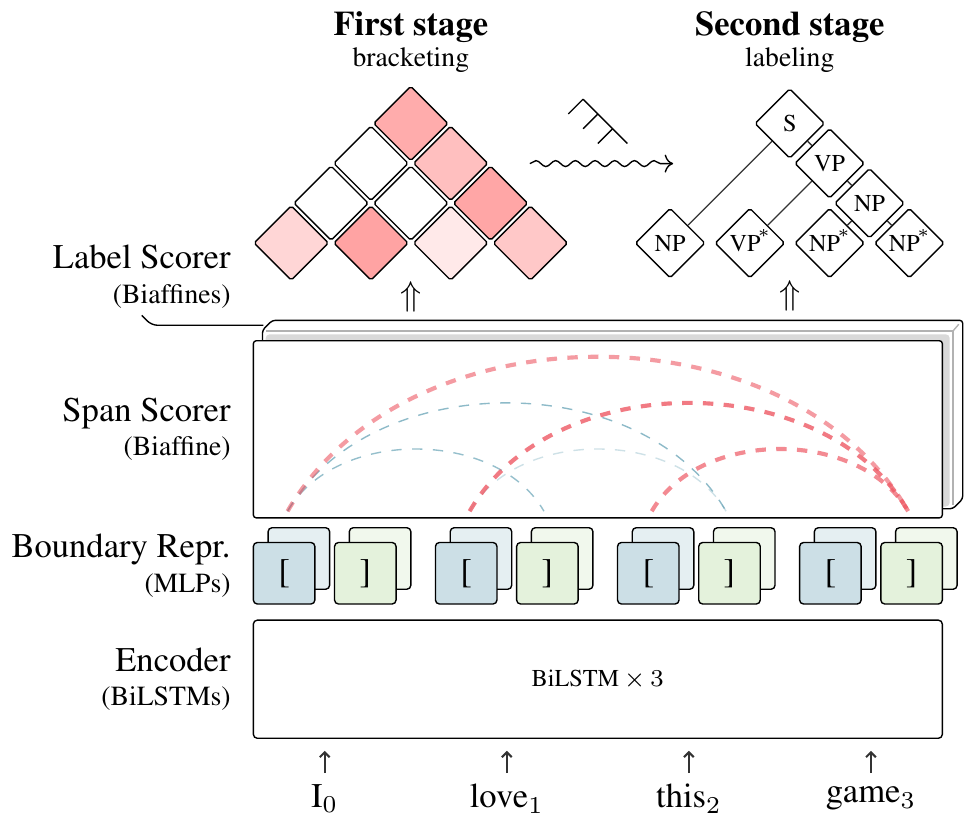}
\caption{Model architecture.
}
\label{fig:framework}
\end{figure}
\section{Two-stage CRF Parsing}
\label{section:2stage-parsing}

Formally, given a sentence consisting of $n$ words  $\boldsymbol{x}=w_0,\dots, w_{n-1}$,
a constituency parse tree, as depicted in Figure~\ref{fig:const-tree-original}, is denoted as $\boldsymbol{t}$, and $(i,j,l) \in \boldsymbol{t}$ is a constituent spanning $w_{i}...w_{j}$ with a syntactic label $l \in \mathcal{L}$.
Alternatively, a tree can be factored into two parts, i.e., $\boldsymbol{t}=(\boldsymbol{y}, \boldsymbol{l})$, where $\boldsymbol{y}$ is an unlabeled (a.k.a. bracketed) tree and $\boldsymbol{l}$ is a label sequence for all constituents in a certain order.
Interchangeably, $(3,5, \texttt{VP})$ is also denoted as $\texttt{VP}_{3,5}$.
% and $(3,5)$ as $UNK_{3,5}$.

% $\boldsymbol{t}=(i,j,l)$, where  can be decomposed into a pair ($\boldsymbol{y}$,$\boldsymbol{l}$), where $\boldsymbol{y}$ is a collection of unlabeled spans:
% $\boldsymbol{y} = \{(i,j):0\le i < n, 0 < j \le n\}$, and each $l \in \mathcal{L}$ in $\boldsymbol{l}$ denotes the label on each span ($i$,$j$).
% (1,4,NP)表示w1...w3对应一个constituent，其label（或non-terminal symbol）是NP。
% POS tags are ignored in the figure since we replace them with the lexical features in this work.

To accommodate the inside and CKY algorithms, we transform the original tree into
%The CRF method only handles the parse trees given in
those of Chomsky normal form (CNF) using the NLTK tool\footnote{\url{https://www.nltk.org}}, as shown in Figure~\ref{fig:binaried-tree}.
Particularly, consecutive unary productions such as $\texttt{X}_{i,j} \rightarrow \texttt{Y}_{i,j}$ are collapsed into one $\texttt{X+Y}_{i,j}$.
We adopt left binarization since preliminary experiments show it is slightly superior to right binarization.
After obtaining the 1-best tree via CKY decoding, the CNF tree is recovered into the \textit{n}-ary form.

% For this reason, we first explicitly binarize each parse tree in the preprocessing process,
% In practice, we found the effect of left-binarization is slightly better than right-binarization.
% The label of dummy inserted span is inherited from its parent and marked with $^{\ast}$ for the post-recovery.

\subsection{Model Definition} \label{sec:model-defination}
\label{sub@section:model-definition}

In this work, we adopt a two-stage bracketing-then-labeling framework for constituency parsing, which we show not only simplifies the model architecture but also improves efficiency, compared with the traditional one-stage approach adopted in previous works \cite{stern-etal-2017-minimal,gaddy-etal-2018-whats}.

%inspired by the recent progress of  dependency parsing \cite{Timothy-d17-biaffine}

% Our parsing approach can be divided into two stages which is one of the main differences between our work and that of pioneers \cite{stern-etal-2017-minimal,gaddy-etal-2018-whats}.
% This helps to simplify the model, improve efficiency, and also leads to slightly better results.

\paragraph{First stage: bracketing.}
Given $\boldsymbol{x}$, the goal of the first stage is to find an optimal unlabeled tree $\boldsymbol{y}$.
The score of a tree is decomposed into the scores of all contained constituents.
\begin{equation} \label{equation:tree-score}
s(\boldsymbol{x},\boldsymbol{y}) = \sum\limits_{(i,j)\in \boldsymbol{y}}s(i,j)
\end{equation}
Under CRF, the conditional probability is
\begin{equation}\label{equation:tree-prob}
\begin{split}
& p(\boldsymbol{y}\mid\boldsymbol{x})  = \frac{e^{s(\boldsymbol{x},\boldsymbol{y})}}{Z(\boldsymbol{x}) \equiv \sum\limits_{\boldsymbol{y'} \in \mathcal{T}(\boldsymbol{x})} {e^{s(\boldsymbol{x},\boldsymbol{y'})}}}
\end{split}
\end{equation}
where $Z(\boldsymbol{x})$ is known as the normalization term, and $\mathcal{T}(\boldsymbol{x})$ is the set of legal trees.

Given all constituent scores $s(i,j)$, we use the CKY algorithm to find the optimal tree $\hat{\boldsymbol{y}}$.
\begin{equation} \label{equation:tree-argmax}
\hat{\boldsymbol{y}} = \arg\max_{\boldsymbol{y}} s(\boldsymbol{x}, \boldsymbol{y}) = \arg\max_{\boldsymbol{y}} p(\boldsymbol{y} \mid \boldsymbol{x})
\end{equation}

% The procedure of searching an optimal unlabeled tree $\hat{\boldsymbol{y}}$ is to find a bracketing tree with the highest conditional probability given the input $\boldsymbol{x}$.

% During test, we use a variant of CKY algorithm to do the tree constraints.
% By excluding the label scores from the bottom-up dynamic programming procedure, the time complexity is reduced to $O(n^3)$ given a sentence of length $n$.

\paragraph{Second stage: labeling.}
Given a sentence $\boldsymbol{x}$ and a tree $\boldsymbol{y}$, the second stage independently predicts a label for each constituent $(i,j) \in \boldsymbol{y}$.
\begin{equation} \label{equation:label-argmax}
\hat{l} = \arg\max_{l \in \mathcal{L}} s(i,j,l)
\end{equation}
Note that we use gold-standard unlabeled trees for loss computation during training.
For a sentence of length $n$, all CNF trees contain the same $2n-1$ constituents. Therefore, this stage has a time complexity of $O(n|\mathcal{L}|)$.

% The next step is to assign labels to all predicted constituents.
% For the predicted constituent, we directly choose the highest-score label $\hat{l}$ and assign to its span ($i$, $j$),
% \begin{equation} \label{equation:label-argmax}
% \hat{l} = \arg\max_{l^{\prime}\in \mathcal{L}} s(i,j,l^{\prime})
% \end{equation}
% forming a well-formed labeled tree $\hat{T}$.

\paragraph{Time complexity analysis.}
The CKY algorithm has a time complexity of $O(n^3)$. Therefore, the time complexity of our two-stage parsing approach is $O(n^3+n|\mathcal{L}|)$.
In contrast, for the one-stage parsing approach, the CKY algorithm needs to determine the best label for all $n^2$ spans and thus needs
$O(n^3+n^2|\mathcal{L}|)$, where $|\mathcal{L}|$ is usually very large (e.g., 138 for English in Table~\ref{table:statistics}).

% Since the number of predicted labels is no more than $n$, the final time complexity is $O(n^3+|\mathcal{L}|n)$, which is further reduced compared with \citeauthor{stern-etal-2017-minimal}~\shortcite{stern-etal-2017-minimal}.

\subsection{Scoring Architecture}

This subsection introduces the network architecture for scoring spans and labels, as shown in Figure~\ref{fig:framework}, which
mostly follows \citeauthor{stern-etal-2017-minimal}~\shortcite{stern-etal-2017-minimal} with two important modifications: 1) boundary representation and biaffine attention for score computation; 2) better parameter settings following \citeauthor{Timothy-d17-biaffine}~\shortcite{Timothy-d17-biaffine}.
% First, we largely borrow the parameter settings from \citeauthor{Timothy-d17-biaffine}~\shortcite{Timothy-d17-biaffine}.
% Second, we use biaffine scoring in place of the minus way \cite{stern-etal-2017-minimal,cross-huang-2016-span,wang-chang-2016-graph}.

\paragraph{Inputs.}
For the $i$th word, its input vector $\mathbf{e}_i$ is the concatenation of the word embedding and character-level representation:
\begin{equation} \label{equation:token-representation}
\mathbf{e}_i = \mathbf{e}^{word}_i \oplus \mathrm{CharLSTM}(w_i)
\end{equation}
where
$\mathrm{CharLSTM}(w_i)$ is the output vectors after feeding the character sequence into a BiLSTM layer \cite{lample-etal-2016-neural}.
Previous works show that replacing POS tag embeddings with $\mathrm{CharLSTM}(w_i)$ leads to consistent improvement \cite{kitaev-klein-2018-constituency}.
This can also simplify the model without the need of predicting POS tags (n-fold jack-knifing on training data).
% Pioneers have shown the usefulness of \mathrm{CharLSTM} \cite{coavoux-crabbe-2017-multilingual}.
% We use it to replace the POS Tagging embeddings and find a consistent improvement.
% The two embeddings are independently dropped before concatenation \cite{Timothy-d17-biaffine}.

\paragraph{BiLSTM encoder.}
We employ three BiLSTM layers over the input vectors for context encoding.
We denote as $\mathbf{f}_i$ and $\mathbf{b}_i$ respectively the output vectors of the top-layer forward and backward LSTMs for $w_i$.

% the top-layer is the concatenation of forward representation $\mathbf{f}_i$ and backward representation $\mathbf{b}_i$.

In this work, we borrow most parameter settings from the dependency parser of \citeauthor{Timothy-d17-biaffine}~\shortcite{Timothy-d17-biaffine}. We find that the dropout strategy is very crucial for parsing performance,
%There are two differences from the
which differs from the vanilla implementation of \citeauthor{stern-etal-2017-minimal}~\shortcite{stern-etal-2017-minimal} in two aspects.

First, for each word $w_i$, $\mathbf{e}^{word}_i$ and $\mathrm{CharLSTM}(w_i)$ are dropped as a whole, either unchanged or becoming a $\mathbf{0}$ vector.
If one vector is dropped into $\mathbf{0}$, the other is compensated with a ratio of 2.
Second, the same LSTM layer shares the same dropout masks at different timesteps (words).
 %are shared the BiLSTM and the MLP in differs from the vanilla implementation in the dropout strategy, which applies the same dropout mask at each timestep.
% We found it crucial for performing competitive performance with \citeauthor{kitaev-klein-2018-constituency}~\shortcite{kitaev-klein-2018-constituency}.
% The two embeddings are independently dropped before concatenation \cite{Timothy-d17-biaffine}.

% For dropouts, we follow \citeauthor{Timothy-d17-biaffine}~\shortcite{Timothy-d17-biaffine} and drop the different components of the input vector $\mathbf{x}_i$  independently.
%Assuming that there are $N$ representations if some are dropped and the rest $M$ ones kept unchanged, we compensate them with a factor of $N/M$. For example, the compensation factor is 2 when there only exists word\&character-level input and one of them is dropped.

\paragraph{Boundary representation.}

For each word $w_i$, we compose the context-aware word representation following \citeauthor{stern-etal-2017-minimal}~\shortcite{stern-etal-2017-minimal}.\footnote{Our preliminary experiments show that $\mathbf{f}_i \oplus \mathbf{b}_{i+1}$ achieves consistent improvement over $\mathbf{f}_i \oplus \mathbf{b}_i$. The possible reason may be that both $\mathbf{f}_i$ and $\mathbf{b}_i$ use $\mathbf{e}_i$ as input and thus provide redundant information.}
\begin{equation}
\mathbf{h}_i = \mathbf{f}_i \oplus \mathbf{b}_{i+1}
\end{equation}
The dimensions of $\mathbf{h}_i$ is 800.

% For the constituent which accommodates a consecutive span ($i$,$j$), $w_i$ and $w_{j-1}$ represent its left and right boundary word respectively.
Instead of directly applying a single MLP to $\mathbf{h}_i$,
we observe that a word must act as either left or right boundaries in all constituents in a given tree.
Therefore, we employ two MLPs to make such distinction and obtain left and right boundary representation vectors.
\begin{equation}
\label{mlp-borlders}
\mathbf{r}_i^{l}; \mathbf{r}_i^{r} =\mathrm{MLP}^{l} \left( \mathbf{h}_i \right); \mathrm{MLP}^{r} \left( \mathbf{h}_i \right)
\end{equation}
The dimension $d$ of $\mathbf{r}_i^{l/r}$ is 500.
As pointed out by \citeauthor{Timothy-d17-biaffine}~\shortcite{Timothy-d17-biaffine}, MLPs reduce the dimension of $\mathbf{h}_i$ and, more importantly, detain only syntax-related information, thus alleviating the risk of over-fitting.

\paragraph{Biaffine scoring.}

Given the boundary representations,
% previous works  often compute the difference vectors and feed them into feedforward layers to compute the span scores.
we score each candidate constituent $(i,j)$ using biaffine operation
%Concretely, we do the attention
over the left boundary representation of $w_i$ and the right boundary representation of $w_j$.
\begin{equation} \label{equation:biaffine}
s(i,j) =  \left[
\begin{array}{c}
  \mathbf{r}_{i}^{l} \\
    1
\end{array}
\right]^\mathrm{T}
\mathbf{W} \mathbf{r}_{j}^{r}
\end{equation}
where $\mathbf{W} \in \mathbb{R}^{d \times d}$.
% The computation regards the span scoring as a boundary matching task, which has somewhat conceptual advantages.

% \paragraph{Label scoring.}
It is analogous to compute scores of constituent labels $s(i,j,l)$.
Two extra MLPs are applied to $\mathbf{h}_i$ to obtain boundary representations $\bar{\mathbf{r}}^{l/r}_i$ (with dimension $\bar{d}$).
We then use $|\mathcal{L}|$ biaffines ($\mathbb{R}^{\bar{d} \times \bar{d}}$) to obtain all label scores.
% of each label $l$ for $(i,j)$.
Since $|\mathcal{L}|$ is large, we use a small dimension $\bar{d}$ of 100 for $\bar{\mathbf{r}}^{l/r}_i$ (vs. 500 for ${\mathbf{r}}^{l/r}_i$) to reduce memory and computation cost.

\paragraph{Previous scoring method.}
\citeauthor{stern-etal-2017-minimal}~\shortcite{stern-etal-2017-minimal} use minus features of BiLSTM outputs as span representations \cite{wang-chang-2016-graph,cross-huang-2016-span} and apply MLPs to compute span scores.
%Basically, they first represent the span ($i$,$j$) by the difference of its two word representations: $\mathbf{r}_{i,j}=\mathbf{r}_i-\mathbf{r}_j$.
%Then, the score of the span with label $l$ is obtained by feeding $\mathbf{r}_{i,j}$ into the feedforward layers with the $\mathrm{ReLU}$ activation:
\begin{equation} \label{equation:minus-score}
s(i,j)=\mathrm{MLP}(\mathbf{h}_{i}-\mathbf{h}_{j})
\end{equation}
We show that our new scoring method is clearly superior in the experiments.

\subsection{Training Loss}

For a training instance $(\boldsymbol{x}, \boldsymbol{y}, \boldsymbol{l})$, The training loss is composed of two parts.
\begin{equation} \label{equation:final-loss}
\mathit{L(\boldsymbol{x}, \boldsymbol{y}, \boldsymbol{l})} = \mathit{L}^{bracket}(\boldsymbol{x}, \boldsymbol{y}) + \mathit{L}^{label}(\boldsymbol{x}, \boldsymbol{y}, \boldsymbol{l})
\end{equation}
The first term is the sentence-level global CRF loss, trying to maximize the conditional probability:
\begin{equation}\label{equation:bracket-loss}
\begin{split}
\mathit{L}^{bracket}(\boldsymbol{x},\boldsymbol{y})
%&=  -\log p(\boldsymbol{y}\mid\boldsymbol{x})  \\
&= -s(\boldsymbol{x}, \boldsymbol{y}) + \log Z(\boldsymbol{x})
\end{split}
\end{equation}
where $\log Z(\boldsymbol{x})$ can be computed using the inside algorithm in $O(n^3)$ time complexity.

The second term is the standard constituent-level cross-entropy loss for the labeling stage.
% By feeding the two loss components together for back-propagation,
% In this work, we use CRF loss in place of the commonly used margin-based training \cite{stern-etal-2017-minimal} for structural learning.
% Another simple local loss is used for the labeled training.

%\paragraph{CRF Loss for spans.}
% The goal of CRF is to maximize the likelihood of the gold parse tree over all possible parse trees.
% Consequently, combined with the definition of the tree probability in Equation~\ref{equation:tree-prob}, the final training loss for brackets becomes
% The first term is the

% \paragraph{Local Loss for labels.}

% Similar to \citeauthor{Timothy-d17-biaffine}~\shortcite{Timothy-d17-biaffine}, we simply employ a cross-entropy loss for label training.
% Given the training instance, the decision of an individual label $l$ is local to the gold-standard span ($i$,$j$) it belongs to, and the label training loss is defined as
% \begin{equation} \label{equation:label-loss}
% \mathit{L}_{label}(i,j) = -s(i,j,l) +\log{\sum_{l' \in \mathcal{L}} e^{s(i,j,l')}}
% \end{equation}

% During training, we optimize the model by directly take the summation of the two losses as the final training objective:

\section{Efficient Training and Decoding}
\label{section:efficient-training-decoding}

\begin{algorithm}[tb]
\caption{Batchified Inside Algorithm.}
\begin{algorithmic}[1]
\newlength{\commentindent}
\setlength{\commentindent}{.2\textwidth}
\renewcommand{\algorithmiccomment}[1]{\unskip\hfill\makebox[\commentindent][l]{$\rhd$~#1}\par}
\LetLtxMacro{\oldalgorithmic}{\algorithmic}
\renewcommand{\algorithmic}[1][0]{%
  \oldalgorithmic[#1]%
  \renewcommand{\ALC@com}[1]{%
  \ifnum\pdfstrcmp{##1}{default}=0\else\algorithmiccomment{##1}\fi}%
}
%\begin{spacing}{1.2}
% \begin{footnotesize}
% \STATE $\forall 0 \le i \le n ~ C_{i, i} = 0$
% \COMMENT $\rhd$ initialization
\STATE \textbf{define:} $S \in \mathbb{R}^{n \times n \times B}$ \COMMENT{$B$ is \#sents in a batch}
% \STATE \hspace{\algorithmicindent}
% \STATE \textbf{Output:} $ C_{0, n} = \log Z(\boldsymbol{x})$
%   \COMMENT{balabala}

\STATE \textbf{initialize:} all $S_{:, :} = 0$
% \STATE $S_{i, i+1} = s_{i, i+1}$ \COMMENT{$w$ is 1}
\FOR [span width]{$w = 1$ \TO $n$}
\STATE \emph{Parallel computation on $0 \le i$,$j<n$,$~r$,$0\le b<B$}
\STATE $S_{i, j=i+w} = \log \sum\limits_{i \le r < j} \exp \left( S_{i, r}+S_{r+1, j} \right)  + s(i, j) $ \label{line:sum-product}\\
  %\textbf{batchify:} $0 \le i$; $j=i+w < n$
\ENDFOR
\RETURN $S_{0, n-1} \equiv \log Z$
% \end{footnotesize}
%\end{spacing}
\end{algorithmic}
\label{alg:inside}
\end{algorithm}

This section describes how we perform efficient training and decoding via batchifying the inside and CKY algorithms for direct computation on GPU. We also show that the very complex outside algorithm can be avoided and fulfilled by the back-propagation process.

% The key to CRF parsing is the efficient computation of the gradients for span scores, more specifically, the inside-outside pass.
% In this section, we propose some initiatives that greatly improve the efficiency of the above procedure, making the CRF parsing feasible and effective.

\subsection{The Batchified Inside Algorithm}

% This work also does the efficiency optimization from the perspective of the inside algorithm itself.
% Unlike the linear-chain CRF, the tree-structure propagation makes the inside algorithm very hard to be batchified.
% This made it unadaptable to the parallel computation on GPU and thus played a continually decreasing role in DL era.
% Previous works often evade this by the Cython programming \cite{kitaev-klein-2018-constituency}.
To compute $\log Z$ in Equation~\ref{equation:bracket-loss} and feature gradients, all previous works on CRF parsing \cite{finkel-etal-2008-efficient,durrett-klein-2015-neural} explicitly perform the inside-outside algorithm on CPUs.
Unlike linear-chain CRF, it seems very difficult to batchify tree-structure algorithms.

% , and
% the major obstacle for solving the efficiency issue  in the DL era is how to
%很多好听的话可以讲：为什么我们能想到

In this work, we find that it is feasible to propose a batchified version of the inside algorithm, as shown in Algorithm~\ref{alg:inside}.
%to adapt to the parallelization on GPU.
The key idea is to pack the scores of same-width spans for all instances in the data batch into large tensors.
This allows us to do computation and aggregation simultaneously via efficient large tensor operation.
%Algorithm~\ref{alg:inside} shows the details.
Since computation for all $0 \le i$,$j<n$,$~r$,$0\le b<B$ is performed in parallel on GPU, the algorithm only needs $O(n)$ steps. Our code will give more technical details.
% The complexity of the inside pass is $O(n)$ parallel steps after batchification.

\subsection{Outside via Back-propagation}

Traditionally, the outside algorithm is considered as indispensable for computing marginal probabilities of subtrees, which further compose feature gradients.
In practice, the outside algorithm is more complex and at least twice slower than the inside algorithm.
Though possible, it is more complicated to batchify the outside algorithm.
Fortunately, this issue is erased in the deep learning era since the back-propagation procedure is designed to obtain gradients. In fact, \citeauthor{eisner-2016-inside}~\shortcite{eisner-2016-inside} proposes a theoretical discussion on the equivalence between the back-propagation and outside procedures.

Since we use a batchified inside algorithm during the forward phase, the back-propagation is conducted based on large tensor computation, which is thus equally efficient.

%which seriously limited the popularity of the CRF-based model in the deep learning era.

% The main bottleneck is the computation of $\log Z(\boldsymbol{x})$ in Equation~\ref{equation:tree-prob} which has a exponential searching space.
% Previous works have solved this by the bottom-up dynamic programming inside algorithm in polynomial time complexity \cite{kasami-1966-efficient, younger-etal-1967-recognition}.

% However, previous non-neural job \cite{finkel-etal-2008-efficient} requires the combination of inside-outside procedure to compute the gradients manually.
% In practice, the outside is twice or more times slower than the inside algorithm, which seriously limited the popularity of the CRF-based model in the deep learning era.
% This work greatly improves the efficiency by replacing the role of outside by the back-propagation.
% We do this replacement thanks for the automatic back-propagation mechanism which is the basic infrastructure in current deep learning library like PyTorch, and theoretical proof of their equivalence \cite{eisner-2016-inside}.

It is also noteworthy that, by setting the loss to $\log Z$ and performing back-propagation, we can obtain the marginal probabilities of spans $(i,j)$, which is exactly the corresponding gradients.
\begin{equation} \label{equation:partial-derivative}
p((i, j)\mid\boldsymbol{x}) = \sum_{\boldsymbol{y}:(i,j) \in \boldsymbol{{y}}} p(\boldsymbol{y}\mid\boldsymbol{x}) = \frac{\partial \log Z(\boldsymbol{x})}{\partial \mathrm{s}(i, j)}
\end{equation}
Marginal probabilities are also useful in many subsequent NLP tasks as soft features.
Please refer to \citeauthor{eisner-2016-inside}~\shortcite{eisner-2016-inside} for more details.

\subsection{Decoding}

As mentioned above, we employ the CKY algorithm to obtain the 1-best tree during the parsing phase, as shown in Equation~\ref{equation:tree-argmax}.
The CKY algorithm is almost identical to the inside algorithm except for replacing the sum-product with a max product (refer to Line~\ref{line:sum-product} in Algorithm~\ref{alg:inside}) and thus can also be efficiently batchified.

% The CKY and inside algorithm actually are very similar, and the difference lies in the use of max-product vs. sum-product (see Line~\ref{ln:inside-sum-product} in Algorithm~\ref{alg:inside}).
% Hence, we batchify the CKY algorithm analogously.
% To ensure the well-formedness of the parsing trees, we first perform a CKY-style decoding algorithm over all spans scores, which rules out the scores of labels.
% The algorithm performs the parsing in a bottom up way to get the predicted unlabeled tree with the highest score:
% \begin{equation}
% \begin{split}
% & \hat{\boldsymbol{y}} = \arg\max_{\boldsymbol{y}} \left[ s(\boldsymbol{x},\boldsymbol{y}) \equiv
% %& s(\boldsymbol{x},\boldsymbol{y}) =
% \sum_{(i, j) \in \boldsymbol{y}}{s(i,j)} \right]  %~~~~ \textit{arc-factorization} \\
% \end{split}
% \end{equation}
% the decoding algorithm is almost identical to the inside algorithm except for replacing the sum product with max product, thus can also be efficiently batchified.
% Then, we directly assign labels in a greedy manner to the corresponding predicted spans.

To perform MBR decoding, we simply replace the span scores $s(i,j)$ with the marginal probabilities $p((i,j)\mid\boldsymbol{x})$ in Equation~\ref{equation:tree-score} and~\ref{equation:tree-argmax}.
%We find and feed them into the decoding algorithm, leading to a slight but consistent improvement.
However, we find this has little influence on parsing performance.%, but reduces the parsing speed due to the need of performing back-propagation for $p((i,j)\mid\boldsymbol{x})$.

\begin{table}[tb]
\centering
\begin{tabular*}{\columnwidth}{@{\extracolsep{\fill}}lrrr|ccc}
\toprule
       & \multirow{2}{*}{\#Train} & \multirow{2}{*}{\#Dev} & \multirow{2}{*}{\#Test} & \multicolumn{2}{c}{\#labels} \\
       & & & & original & CNF \\[1pt]
       \midrule
       % \\[-8pt]
PTB    &  39,832 & 1,700 &  2,416 & 26 &  138 \\
CTB5.1 &  18,104 &   352 &    348 & 26 &  162 \\
CTB7   &  46,572 & 2,079 &  2,796 & 28 &  265 \\
\bottomrule
\end{tabular*}
\caption{Data statistics, including the number of sentences and constituent labels. For ``\#labels'', we list the number of labels in both original and converted CNF trees.}
\label{table:statistics}
\end{table}
\begin{table*}[tb]
\centering
\begin{tabular*}{\textwidth}{@{\extracolsep{\fill}}lccccccccccc}
\toprule
% & \multicolumn{3}{c}{Unlabeled} & \multicolumn{3}{c}{Labeled} \\
% & P & R & F & P & R & F \\
% Max-margin (minus) &         94.47  &         94.41  &         94.44  &         93.54  &         93.47  &         93.51  \\
% Max-margin         &         94.63  &         94.66  &         94.65  &         93.70  &         93.73  &         93.72  \\
% CRF (mix training) &         94.27  &         94.54  &         94.40  &         93.35  &         93.62  &         93.49  \\
% CRF w/o MBR       &         94.65  &         94.76  &         94.70  &         93.75  &         93.85  &         93.80  \\
% CRF                & \textbf{94.67} & \textbf{94.87} & \textbf{94.77} & \textbf{93.77} & \textbf{93.96} & \textbf{93.86} \\
& \multicolumn{3}{c}{PTB} && \multicolumn{3}{c}{CTB5.1} && \multicolumn{3}{c}{CTB7} \\
% \cmidrule(lr){2-4}\cmidrule(lr){6-8}\cmidrule(lr){10-12}
& P & R & F && P & R & F && P & R & F \\
\midrule
% \\[-8pt]
Max-margin (one-stage) &         93.70  &         93.73  &         93.72  &&         90.60  &         90.48  &         90.54  &&         86.85  &         86.08  &         86.47  \\
CRF (one-stage)        &         93.44  &         93.75  &         93.60  && \textbf{91.08} &         90.98  & \textbf{91.03} &&         87.10  &         86.75  &         86.93  \\[3pt]
CRF (two-stage)        & \textbf{93.77} & \textbf{93.96} & \textbf{93.86} &&         90.91  &         91.09  &         91.00  && \textbf{87.27} & \textbf{87.00} & \textbf{87.13} \\
\qquad w/o MBR         &         93.75 &          93.85  &         93.80  &&         90.93  & \textbf{91.10} &         91.02  &&         87.21  &         86.89  &         87.05  \\
% \hline
\qquad minus features  &         93.40  &         93.35  &         93.37  &&         90.60  &         90.51  &         90.56  &&         86.96  &         86.24  &         86.60  \\
\qquad vanilla dropout &         92.80  &         93.00  &         92.90  &&         89.68  &         89.68  &         89.68  &&         85.55  &         85.54  &         85.54  \\

\bottomrule
\end{tabular*}
\caption{Results on dev data. All models use randomly initialized word embeddings.}
\label{table:dev}
\end{table*}

\section{Experiments}
\label{section:experiments}
\paragraph{Data.}
We conduct experiments on three English and Chinese datasets.
The first two datasets, i.e., PTB and CTB5.1, are widely used in the community.
%i.e., the English PTB (WSJ) and the Chinese CTB v5.1.
We follow the conventional train/dev/test data split.
% The training data are selected from Section 02-21 while dev and test data are from Section 22 and 23 respectively.
% For Chinese, we adopt the CTB5.1  and follow the settings of \citeauthor{liu-zhang-2017-shift}~\shortcite{liu-zhang-2017-shift}.
% That is, articles 001-270 and 440-1151 are used for training, articles 301-325, and 271-300 for dev and test severally.
Considering that both CTB5.1-dev/test only have about 350 sentences,
% we also use the larger CTB7 for more robust investigations.
we also use the larger CTB7 for more robust investigations, following the data split suggested in the official manual.
%We also compare the models on CTB7,
%a newer and large-scale version, to further examine the impact of the CRF approach.
% We follow the data split suggested in the official manual.
% Note that, CTB5.1 is a kind of small dataset consisting of roughly 19 thousand sentences. Both dev and test set only include half of a thousand sentences.
% It is not good enough to probe a deep neural network.
% Hence we also evaluate our approach in version 7.0 of Penn Chinese Treebank (CTB7) which has 46 thousand sentences for training and at least two thousand sentences in both dev and test set.
% \textcolor{red}{Talk about the splitting detail of CTB7.}
Table~\ref{table:statistics} shows the data statistics.
We can see that CNF introduces many new constituent label types,  most (about 75\%) of which are from the collapsing process of consecutive unary rules.

%(see discussion) as discussed in Section \ref{section:2stage-parsing}.
% Please kindly note that most of the added labels come from the collapse-unary process.
% Following \citeauthor{gaddy-etal-2018-whats}~\shortcite{gaddy-etal-2018-whats}, low-frequency labels are not filtered out.

% \paragraph{Data preprocessing.}
% Following standard conventions, we remove all ``-NONE-'' label from both of Chinese Treebank.
% For our approach requests explicit binarization, we using NLTK package\footnote{https://www.nltk.org/} to left binarize our dataset into Chomsky Normal Form (CNF) then we collapse all the unary chains.
% Note that we do binarization before collapsing to avoid the explosion of the label.

\paragraph{Evaluation.}
%We need a post-recovery process to convert (debinarize) the parsed trees to the form in Figure~\ref{fig:const-tree-original} before evaluation.
As mentioned earlier, we convert 1-best CNF trees into \textit{n}-ary trees after parsing for evaluation.
Here, it may be useful to mention a small detail.
The predicted 1-best CNF tree may contain inconsistent productions since the decoding algorithm does not have such constraints.
Taking Figure~\ref{fig:binaried-tree} as an example, the model may output $\texttt{VP}_{3,5} \rightarrow \texttt{PP}^{\ast}_{3,3} ~ \texttt{NP}_{4,5}$, where $\texttt{VP}$ is incompatible with $\texttt{PP}^{\ast}$.
During the \textit{n}-ary post-processing, we simply ignore the concrete label string $\texttt{PP}$ before the ``$\mathtt{\ast}$'' symbol.
In view of this, performance may be slightly improved by adding such constraints during decoding.

We use the standard constituent-level labeled precision, recall, F-score (P/R/F) as the evaluation metrics with the \texttt{EVALB} tool\footnote{\url{https://nlp.cs.nyu.edu/evalb}}.
%P/R represents the proportion of correct labeled constituents in the golden/parsed instances.
Specifically, a predicted constituent such as $\texttt{VP}_{3,5}$ is considered correct if it also appears in the gold-standard tree.\footnote{
Since some researchers may implement their own evaluation scripts, some details about \texttt{EVALB} need to be clarified for fair comparison:
1) Empty constituents like \{-NONE-\} are removed during data pre-processing.
2) Root constituents (\{TOP, S1\} for English and an empty string for Chinese) are ignored for evaluation.
3) Constituents spanning a English punctuation mark like \{:, ``, '', ., ?, !\} are also ignored. Please note that Chinese punctuation marks are evaluated as normal words.
4) Some label sets like \{ADVP, PRT\} are regarded as equivalent.}
% By default, some special labels\footnote{
% The root labels \{TOP, S1\} and punctuations like \{:, ``, '', ., ?, !\} are ignored.
% Some placeholders like \{-NONE-\} are also stripped away.
% }
% are excluded from the evaluation and some labels like ``ADVP'' and ``PRT'' are regarded as equivalent.

% eval：第一句：把CNF还原成图1的形式；
% 第二句：用evalb评价(不要提重现)；
% 第三句：稍微介绍一下LR LP LF的含义，和标点的情况。
% We estimate the performance of our models using EVALB\footnote{https://nlp.cs.nyu.edu/evalb/} before we uncollapse unary chains and debinarize parsing trees.
% To avoid the high-frequency files reading and writing during training, we also re-implement a faster python version evaluation script\footnote{url:} yielding the same scores as EVALB do.
% Following standard conventions, we remove all ``-NONE-'' label and functional labels from all treebanks, treat ``ADVP'' and ``PRT'' as equal, and delete punctuation labels when evaluating English treebanks.
% Then we measure the model performance in terms of Labeled Precision (P), Labeled Recall (R) and Labeled $\text{F}_1$ scores (F).

\paragraph{Parameter settings.}
% 描述一下stern是怎么做的解释一下和我们的区别(不重要的参数不用那么啰嗦)
% Dropout那一块很重要，一定要说清楚。
% \textcolor{red}{正常的dropout：embedding部分先将两个向量拼起来再dropout，而不是先各自dropout再拼起来。LSTM部分：每个timestep的dropout用不同的mask将元素置零。}
% Instead of tuning the hyper-parameters elaborately, we directly borrow most of the  settings from \citeauthor{Timothy-d17-biaffine}~\shortcite{Timothy-d17-biaffine}.
We directly adopt the same hyper-parameter settings of the dependency parser of \citeauthor{Timothy-d17-biaffine}~\shortcite{Timothy-d17-biaffine} without further tuning.
The only difference is the use of CharLSTM word representations instead of POS tag embeddings.
The dimensions of char embedding, word embedding, and CharLSTM outputs are
50, 100, 100, respectively.
All dropout ratios are 0.33.
The mini-batch size is 5,000 words.
% For pretrained word embeddings, we adopt the English GloVe embeddings\footnote{\url{https://nlp.stanford.edu/projects/glove/}}, and train word embeddings on Chinese Gigaword Third Edition using word2vec.
The training process continues at most 1,000 epochs and is stopped if the peak performance on dev data does not increase in 100 consecutive epochs.
% The dimensions of word embeddings and CharLSTM outputs is set to 100.

% Specifically, the dimension of word embeddings are 100 and character embeddings are 50, both of them are random initialized by the default initialization of embedding layer in PyTorch.
% The hidden size of the one layer CharLSTM is 100.
% We then encode sentences using a three-layer 400-dimensional BiLSTM whose weights are initialized with an orthogonal matrix while biases are zero-initialized.
% The hidden size of BiLSTM outputs is 800, and MLPs extract the representations with the size of 500 and 100 for spans and labels respectively.
% We initialize the MLPs by the default kaiming normal initialization of PyTorch.
% As for biaffine layers, we fill the parameter matrixes with values drawn from $\mathcal{N}\left(0, 1^2\right)$.

% During training, we set our mini-batch to a collection of sentences with approximate 5000 words and all dropout probabilities to 0.33.
% The training process continues at most 1,000 epochs and is stopped if the peak performance on dev data does not increase in 100 consecutive epochs.
% \% and update our model's parameters using an annealed Adam optimizer.
% All parameters of the optimizer are also identical to \citeauthor{Timothy-d17-biaffine}~\shortcite{Timothy-d17-biaffine}.
% we then batch sentences together by approximate length.
% Each batch contains a set of sentences consisting of 5000 words.

%\subsection{Results on Dev Data}

\subsection{Model Comparison on Dev Data}

We conduct the model study on dev data from two aspects: 1) CRF vs. max-margin training loss; 2) two-stage vs. one-stage parsing.
The first three lines of
Table~\ref{table:dev} shows the results.
The three models use the same scoring architecture and parameters.
Following previous practice \cite{stern-etal-2017-minimal}, one-stage models use only scores of labeled constituents $s(i,j,l)$.
%on the dev data, including the max-margin and our CRF approach.
% \footnote{We perform MBR decoding on all datasets for the CRF approach and the improvement brought by MBR is very slight. We ignore the exhibition of the results without MBR due to space limitation.}
% For fairness, we retain the biaffine scoring in the max-margin method.
In order to verify the effectiveness of the two-stage parsing, we also list the results of ``CRF (one-stage)'', which directly scores labeled constituents.
\begin{equation} \label{equation:tree-label-score}
s(\boldsymbol{x},\boldsymbol{y},\boldsymbol{l}) =
%\sum_{(i,j)\in \boldsymbol{y}} {s(i,j) +
\sum_{(i,j,l) \in (\boldsymbol{y}, \boldsymbol{l})} s(i,j,l)
\end{equation}
As discussed in the last paragraph of Section \ref{sec:model-defination}, the inside and CKY algorithms become a bit more complicated for the one-stage parser that two-stage.

From the first two rows, we can see that
under the one-stage parsing framework, the CRF loss leads to similar performance on English
but consistently outperforms the max-margin loss by about 0.5 F-score on both Chinese datasets.
The max-margin loss has one extra hyper-parameter, namely the margin value, which is set to 1 according to preliminary results on English and not tuned on Chinese for simplicity.
We suspect that the performance on Chinese with max-margin loss may be improved with more tuning.
%First, we compare the one-stage CRF approach, with widely used max-margin loss.
%The results of max-margin are very close to that of CRF in English.
%But on Chinese, CRF outperforms the max-margin consistently by about 0.5.
%This is partly because we do not tune hyper-parameters for max-margin specifically.
%We believe that max-margin still has the potential to catch up with our CRF approach with a proper setting.
% the CRF parsing span and label together is named as ``CRF w/ label''. its architecture is identical to ``Max-margin'' but loss function.
Overall, we can conclude that the two training loss settings achieve very close performance, and CRF has an extra advantage of probabilistic modeling.

Comparing the second and third rows, the two CRF parsers achieve nearly the same performance on CTB5.1 and the two-stage parser achieves modest improvement over the one-stage parser by about 0.2 F-score on both PTB and CTB7.
% We can see that the models adopted CRF-based loss outperform those with max-margin loss in most data.
% The only exception is the CRF loss for labels results in significant degradation of performance on PTB.
% It may because the masses of rare labels do not cooperate well with CRF loss function.
%Secondly, we list the comparison between one- and two-stage CRF\footnote{Unlike CRF, max-margin extra incorporates the hamming loss into labeled spans and the labeled span score $s(i,j,l)$ becomes $s(i,j,l)+1(l^{\ast}_{i,j}\neq l_{i,j})$ \cite{stern-etal-2017-minimal}. Thus max-margin is unsuited to be factorized into two-stage training.}.
% The results are very similar.
%Concretely, ``CRF (two-stage)'' achieves a slight improvement of 0.2 on PTB and CTB7 over ``CRF (one-stage)'', and underperforms by 0.03 on CTB5.1.
% This indicates that our separate training approach can improve the parsing efficiency without hurting performance.
% the model produces a great improvement on PTB and becomes superior to ``Max-margin''.
% Besides, separated training also yields competitive performances in Chinese data.
% Especially, it outperforms the unseparated one by 0.2 F on CTB7.
Therefore, we can conclude that our proposed two-stage parsing approach is superior in simplicity and efficiency (see Table~\ref{table:speed}) without hurting performance.

\begin{table}[tb]
\centering
\begin{tabular*}{\columnwidth}{@{\extracolsep{\fill}}lr}
\toprule
& Sents/sec \\
\midrule
% \\[-8pt]
\citeauthor{petrov-klein-2007-improved}~\shortcite{petrov-klein-2007-improved}  (Berkeley Parser)          & 6   \\
\citeauthor{zhu-etal-2013-fast}~\shortcite{zhu-etal-2013-fast} (ZPar)                                      & 90  \\
\citeauthor{stern-etal-2017-minimal}~\shortcite{stern-etal-2017-minimal}                                   & 76  \\
\citeauthor{shen-etal-2018-straight}~\shortcite{shen-etal-2018-straight}                                   & 111 \\
% \citeauthor{zhou-zhao-2019-head}~\shortcite{zhou-zhao-2019-head}                                          & 159 \\
\citeauthor{kitaev-klein-2018-constituency}~\shortcite{kitaev-klein-2018-constituency}                     & 332 \\
\citeauthor{gomez-rodriguez-vilares-2018-constituent}~\shortcite{gomez-rodriguez-vilares-2018-constituent} & 780 \\[3pt]
% Max margin (Cython)                                                & 617 \\
% Max margin                                                         & 1056 \\
% labeled CRF (Cython)                                               & 369 \\
% labeled CRF w/o MBR                                                & 990 \\
% labeled CRF                                                        & 617 \\
% CRF (Cython)                                                       & 722 \\
CRF (one-stage)                                                    & 990 \\
CRF (two-stage) w/ MBR                                             & 743 \\
CRF (two-stage) w/o MBR                                            & \textbf{1092} \\

\bottomrule
\end{tabular*}
\caption{Speed comparison on PTB test.
% $\textrm{Ours}^\sharp$ indicates the CRF model with MBR decoding.
}
\label{table:speed}
\end{table}
\begin{table*}[tb]
\centering
\begin{tabular*}{\textwidth}{@{\extracolsep{\fill}}lccccccccccc}
\toprule
& \multicolumn{3}{c}{PTB} && \multicolumn{3}{c}{CTB5.1} && \multicolumn{3}{c}{CTB7} \\
& P & R & F && P & R & F && P & R & F \\
\midrule
% \\[-8pt]
% \multicolumn{4}{l}{\textbf{Random word embeddings}} \\
% \citeauthor{liu-zhang-2017-shift}~\shortcite{liu-zhang-2017-shift}                     &         92.1\textcolor{white}{0}   &         91.3\textcolor{white}{0}   &         91.7\textcolor{white}{0}   &&         85.9\textcolor{white}{0}   &         85.2\textcolor{white}{0}   &         85.5\textcolor{white}{0}   &&         -      & -              & -              \\
\citeauthor{stern-etal-2017-minimal}~\shortcite{stern-etal-2017-minimal}                  &         92.98  &        90.63   &         91.79  &&         -      &         -      &         -      &&         -      & -              & -              \\
% \citeauthor{liu-2018-improving}~\shortcite{liu-2018-improving}                       &         -      &         -      &         91.2   &&         -      &         -      &         84.1   &&         -      & -              & -              \\
% \citeauthor{fried-klein-2018-policy}~\shortcite{fried-klein-2018-policy}                       &         -      &         -      &         92.2   &&         -      &         -      &         87.0   &&         -      & -              & -              \\
% \citeauthor{stern-etal-2017-effective}~\shortcite{stern-etal-2017-effective}                &         92.57   &          92.56   &         92.56  \\
\citeauthor{gaddy-etal-2018-whats}~\shortcite{gaddy-etal-2018-whats}                    &         92.41    &         91.76    &         92.08    &&         -        &         -        &         -        &&          -     &         -      &         -      \\
\citeauthor{kitaev-klein-2018-constituency}~\shortcite{kitaev-klein-2018-constituency}           &         93.90    &         93.20    &         93.55    &&         88.09    &       86.78      &         87.43    &&          -     &         -      &         -      \\
\citeauthor{gomez-rodriguez-vilares-2018-constituent}~\shortcite{gomez-rodriguez-vilares-2018-constituent} &         -        &         -        &         90.0\textcolor{white}{0} &&         -        &         -        &         84.4\textcolor{white}{0} &&          -     &         -      &         -      \\
\citeauthor{shen-etal-2018-straight}~\shortcite{shen-etal-2018-straight}                  &         92.0\textcolor{white}{0} &         91.7\textcolor{white}{0} &         91.8\textcolor{white}{0} &&         86.6\textcolor{white}{0} &         86.4\textcolor{white}{0} &         86.5\textcolor{white}{0} &&          -     &         -      &         -      \\
\citeauthor{teng-zhang-2018-two}~\shortcite{teng-zhang-2018-two} %(w/ pretrained)
                                                 &         92.5\textcolor{white}{0} &         92.2\textcolor{white}{0} &         92.4\textcolor{white}{0} &&         87.5\textcolor{white}{0} &         87.1\textcolor{white}{0} &         87.3\textcolor{white}{0} &&         -      &         -      & -    \\

\citeauthor{vilares-etal-2019-better}~\shortcite{vilares-etal-2019-better}                 &         -        &         -        &         90.60    &&         -        &         -        &         85.61    &&          -     &         -      &         -      \\
\citeauthor{zhou-zhao-2019-head}~\shortcite{zhou-zhao-2019-head} w/ pretrained                        &         93.92                    &         93.64  &         93.78  &&    89.70 & 89.09 & 89.40        &&         -      &         -      & -    \\[3pt]
% Ours w/o MBR                                   & \textbf{93.82} & \textbf{93.53} & \textbf{93.67} \\
% Ours                                             & \textbf{93.84} & \textbf{93.58} & \textbf{93.71} && \textbf{89.18} & \textbf{89.03} & \textbf{89.10} && \textbf{87.66} &         \textbf{87.21}  &         \textbf{87.43}  \\
Ours                              &         93.84    &         93.58    &         93.71    &&         89.18    &         89.03    &         89.10    &&         87.66  &         87.21  &         87.43  \\

% \multicolumn{4}{l}{\textbf{Pretrained word embeddings}} \\
Ours w/ pretrained                                             & \textbf{94.23}   & \textbf{94.02}   & \textbf{94.12}   && \textbf{89.71}   & \textbf{89.89}   & \textbf{89.80}   && \textbf{88.84} & \textbf{88.36} & \textbf{88.60} \\[1pt]
% \multicolumn{4}{l}{\textbf{ELMo/BERT}} \\
\hline
\\[-8pt]
\citeauthor{kitaev-klein-2018-constituency}~\shortcite{kitaev-klein-2018-constituency} w/ ELMo & 95.40 & 94.85 & 95.13 && - & - & - && - & -& - \\
\citeauthor{kitaev-etal-2019-multilingual}~\shortcite{kitaev-etal-2019-multilingual} w/ BERT           &         95.73  &    95.46       &         95.59  &&         91.96  &     91.55      &         91.75  &&         -      &         -      &         -      \\[3pt]
%\citeauthor{zhou-zhao-2019-head}~\shortcite{zhou-zhao-2019-head}                      &         95.70  & \textbf{95.98}  & \textbf{95.84} w/ Large BERT?  &&         -      &         -      &         -      &&         -      &         -      &         -      \\
% Ours (Max margin) + Pretrained & 94.08 & 93.72 & 93.90 \\
Ours w/ BERT                                             & \textbf{95.85} &         \textbf{95.53}  &         \textbf{95.69}  && \textbf{92.51} & \textbf{92.04} & \textbf{92.27} && \textbf{91.73} & \textbf{91.38} & \textbf{91.55} \\
\bottomrule
\end{tabular*}
\caption{Results on test data.}
\label{table:test}
\end{table*}

\subsection{Ablation Study on Dev Data}

To gain insights into the contributions of individual components in our proposed framework,
% mainly contributes to the power of our proposed CRF framework,
we then conduct the ablation study by undoing one component at a time. % to \citeauthor{stern-etal-2017-minimal}~\shortcite{stern-etal-2017-minimal}.
Results are shown in the bottom four rows of Table~\ref{table:dev}.

\paragraph{Impact of MBR decoding.}
By default, we employ CKY decoding over marginal probabilities, a.k.a. MBR decoding.
The ``w/o MBR'' row presents the results of performing decoding over span scores.
%the CRF parser that  employing the MBR decoding.
Such comparison is very interesting since it is usually assumed that MBR decoding is theoretically superior to vanilla decoding.
However, the results clearly show that
the two decoding methods achieve nearly identical performance.
%The overall performance is almost the same as our main CRF results.
% From this, we can conclude that, using marginal probabilities for decoding has very slight impact on the final results.
% First, We withdraw MBR decoding from our framework.

\paragraph{Impact of scoring architectures.}
In order to measure the effectiveness of our new scoring architecture, we revert the biaffine scorers to the ``minus features'' method adopted by \citeauthor{stern-etal-2017-minimal}~\shortcite{stern-etal-2017-minimal} (refer to Equation~\ref{equation:minus-score}).
% Basically, they first represent the span ($i$,$j$) by the difference of its two word representations: $\mathbf{r}_{i,j}=\mathbf{r}_i-\mathbf{r}_j$.
% Then, the score of the span with label $l$ is obtained by feeding $\mathbf{r}_{i,j}$ into the feedforward layers with the $\mathrm{ReLU}$ activation:
% \begin{equation} \label{equation:minus-score}
% s(i,j,l)=[\mathbf{W}_2\mathrm{ReLU}(\mathbf{W}_1 \mathbf{r}_{i,j}+\mathbf{b}_1)+\mathbf{b}_2]_l
% \end{equation}
% At first, we replace our biaffine scorer with original minus-feature which are used in many recent work\cite{cross-huang-2016-span,stern-etal-2017-minimal,gaddy-etal-2018-whats,kitaev-klein-2018-constituency,kitaev-etal-2019-multilingual,zhou-zhao-2019-head}.
It is clear that our proposed scoring method is superior to the widely used minus-feature method, and
achieves a consistent and substantial improvement of about 0.5 F-score on all three datasets.
%With this replacement, ``minus-feature'' lags behind our main results by 0.49, 0.44 and 0.53 on PTB, CTB5.1, and CTB7 respectively.
% We can conclude that, our biaffine  is a more effective way of span scoring compared with the commonly used minus-feature.
% The results clearly prove the superiority of our biaffine scoring mechanism.

\paragraph{Impact of dropout strategy.}
We keep other model settings unchanged and only replace the dropout strategy borrowed from \citeauthor{Timothy-d17-biaffine}~\shortcite{Timothy-d17-biaffine} with the vanilla dropout strategy adopted by \citeauthor{stern-etal-2017-minimal}~\shortcite{stern-etal-2017-minimal}.
This leads to a very large and consistent performance drop of 0.96, 1.39 and 1.59 in F-score on the three datasets, respectively.
\citeauthor{kitaev-klein-2018-constituency}~\shortcite{kitaev-klein-2018-constituency} replaced BiLSTMs with a self-attention encoder in \citeauthor{stern-etal-2017-minimal}~\shortcite{stern-etal-2017-minimal} and achieved a large improvement of 1.0 F-score by separating content and position attention.
Similarly, this work shows that the BiLSTM-based parser can be very competitive with proper parameter settings.

\subsection{Speed Comparison}
Table~\ref{table:speed} compares different parsing models in terms of parsing speed.
%shows the parsing speed of our model and previous representative parsers.
% We list the parsing speed of our methods and prior research on Table~\ref{table:speed}.x
%For fair comparison, all the speed of
Our models are both run on a machine with Intel Xeon E5-2650 v4 CPU and Nvidia GeForce GTX 1080 Ti GPU.
% All other numbers are obtained from the corresponding papers.
% Note that, \citeauthor{kitaev-klein-2018-constituency}~\shortcite{kitaev-klein-2018-constituency} list no parsing speed in their paper.
% Thus we also measure its parsing speed in the same machine.
% Labeled CRF means the model does not separate the span and label training.
% The notation Cython refers to those models using the same Cython version of cky-style decoding function implemented by \citeauthor{kitaev-klein-2018-constituency}~\shortcite{kitaev-klein-2018-constituency}.
Berkeley Parser and ZPar are two representative non-neural parsers without access to GPU.
\citeauthor{stern-etal-2017-minimal}~\shortcite{stern-etal-2017-minimal} employ max-margin training and perform CKY-like decoding on CPUs.
\citeauthor{kitaev-klein-2018-constituency}~\shortcite{kitaev-klein-2018-constituency} use a self-attention encoder and perform decoding using Cython for acceleration.

We can see that our one-stage CRF parser is much more efficient than previous parsers by directly performing decoding on GPU.
Our two-stage parser can parse 1,092 sentences per sentence, which is three times faster than \citeauthor{kitaev-klein-2018-constituency}~\shortcite{kitaev-klein-2018-constituency}.
Of course, it is noteworthy that those parsers \cite{stern-etal-2017-minimal,kitaev-klein-2018-constituency} may be equally efficient by adopting our batchifying techniques.

The parser of \citeauthor{gomez-rodriguez-vilares-2018-constituent}~\shortcite{gomez-rodriguez-vilares-2018-constituent} is also very efficient by treating parsing as a sequence labeling task. However, the parsing performance is much lower, as shown in Table~\ref{table:test}.

The two-stage parser is only about 10\% faster than the one-stage counterpart. The gap seems small considering the significant difference in time complexity as discussed (see Section~\ref{sub@section:model-definition}).
The reason is that the two parsers share the same encoding and scoring components, which consume a large portion of the parsing time.

Using MBR decoding requires an extra run of the inside and back-propagation algorithms for computing marginal probabilities, and thus is less efficient.
As shown in Table~\ref{table:dev}, the performance gap is very slight between w/ and w/o MBR.

\subsection{Results and Comparison on Test Data}
%The results of English and Chinese on the test data is shown in
Table~\ref{table:test} shows the final results on the test datasets under two settings, i.e., w/o and w/ ELMo/BERT.

Most previous works do not use pretrained word embedding but use randomly initialized ones instead, except for \citeauthor{zhou-zhao-2019-head}~\shortcite{zhou-zhao-2019-head}, who use Glove for English and structured skip-gram embeddings.
For pretrained word embeddings, we use Glove (100d) %\cite{pennington-etal-2014-glove}
for English PTB\footnote{\url{https://nlp.stanford.edu/projects/glove}},
and adopt the embeddings of \citeauthor{li-etal-2019-attentive}~\shortcite{li-etal-2019-attentive} trained on Gigaword 3rd Edition for Chinese.
% Looking at the results of using pretrained word embeddings,
It is clear that our parser benefits substantially from the pretrained word embeddings.\footnote{
We have also tried the structured skip-gram embeddings kindly shared by \citeauthor{zhou-zhao-2019-head}~\shortcite{zhou-zhao-2019-head} for Chinese, and achieved similar performance by using our own embeddings.
}

We also make comparisons with recent related works on constituency parsing, as discussed in Section~\ref{section:relwork}.
% As listed in Table~\ref{table:dev}, employing MBR decoding performs slightly better than directly decoding with scores.
% Based on this observation,
% we report the values with MBR decoding as our final results.
% Based on the results on dev data, we choose the highest performance CRF model to compare with previous work.
% We also measure the performance of the self-attention model \citeauthor{kitaev-klein-2018-constituency}~\shortcite{kitaev-klein-2018-constituency} on both CTB5.1 and CTB7 by their open-source code.
We can see that our BiLSTM-based parser outperforms the basic \citeauthor{stern-etal-2017-minimal}~\shortcite{stern-etal-2017-minimal} by a very large margin, mostly owing to the new scoring architecture and better dropout settings.
%In the scenario of using random initialized word embeddings, our parser outperforms prior works which use BiLSTM encoder and max-margin loss \cite{stern-etal-2017-minimal,gaddy-etal-2018-whats} by very large margin.
Compared with the previous state-of-the-art self-attentive parser \cite{kitaev-klein-2018-constituency},
our parser achieves an absolute improvement of 0.16 on PTB and 1.67 on CTB5.1 without any language-specific settings.
% According to our knowledge, the K\&K parser used different parameter settings for the two languages.
% Compared with those parsers using random word embeddings, we can see that our CRF model achieves the best performance on both PTB and CTB5.1 but CTB7.
% The self-attention model surpasses ours by 0.19 F on CTB7 for having more parameters.
% Once we increase the number of parameters of our model to a similar level, enlarge the dimension of character embeddings to 64 and the layer number of BiLSTM to 4, our model achieved better performance, 87.69 F on CTB7, and be superior to the self-attention model.

% We then utilize the extra resources.
% The results with pretrained word embeddings are also reported for comparison.
% First, we incorporate pretrained word embeddings into our CRF model.
% It is noteworthy that our CRF model using pretrained word embedding outperforms pretrained version of the state-of-the-art HPSG parser \cite{zhou-zhao-2019-head}, which regards constituency and dependency parsing as a special kind of multi-task learning problem, by 0.34 F on PTB.
% 1）他们论文中的结果错误的使用了正确词性。我们重跑他们的代码，用正确词性的结果是92.xxx，和他们论文汇报的结果一致。我们的parser，用正确词性，可以从89.80提升到92.xx。
% 2）自动词性是怎么产生的。
% 3）structured skip-gram和我们的embedding性能接近。
% 4）Anoter detail need to be clarified: data split and Stanford dependencies 3.3.0.

The CTB5.1 results of \citeauthor{zhou-zhao-2019-head}~\shortcite{zhou-zhao-2019-head} is obtained by rerunning their released code using predicted POS tags.
We follow their descriptions\footnote{\url{https://github.com/DoodleJZ/HPSG-Neural-Parser}} to produce the POS tags.
% predicted by the Stanford tagger \cite{toutanova-etal-2003-feature}.
It is noteworthy that their reported results accidentally use gold POS tags on CTB5.1, which is confirmed after several turns of email communication. We are grateful for their patience and help.
We reran their released code using gold POS tags, and got 92.14 in F-score on CTB5-test, very close to the results reported in their paper.
Our parser achieves 92.66 F-score with gold POS tags.
Another detail about their paper should be clarified: for dependency parsing on Chinese, they adopt two different data split settings, both using Stanford dependencies 3.3.0 and gold POS tags.

The bottom three rows list the results under the setting of using ELMo/BERT.
We use bert-large-cased\footnote{\url{https://github.com/huggingface/transformers}} (24 layers, 1024 dimensions, 16 heads) for PTB following \citeauthor{kitaev-etal-2019-multilingual}~\shortcite{kitaev-etal-2019-multilingual}, and bert-base-chinese (12 layers, 768 dimensions, 12 heads) for CTB.
It is clear that using BERT representations can help our parser by a very large margin on all datasets. %, which is consistent with previous works.
Our parser also outperforms the multilingual parser of \citeauthor{kitaev-etal-2019-multilingual}~\shortcite{kitaev-etal-2019-multilingual}, which uses extra multilingual resources.
In summary, we can conclude that our parser achieves state-of-the-art performance in both languages and both settings.

% Finally, we report the results using ELMo \cite{peters-etal-2018-deep} and BERT \cite{devlin-etal-2019-bert}.
% When taking the same BERT settings, our results outperform \citeauthor{kitaev-etal-2019-multilingual}~\shortcite{kitaev-etal-2019-multilingual} which also introduce multilingual resources, and achieve the state-of-the-art.
% Secondly, We experiment with BERT, a contextualized word representation.
% Our CRF model with BERT slightly underperforms the BERT version HPSG parser but still surpasses the performances of other single-task models.

\section{Related Works}
\label{section:relwork}

% Due to space limitation, we try to confine our discussions on neural constituency parsing, most of which are compared in
%We make comparisons with most related works in
% Table~\ref{table:test}.

% The two most closely related works are \citeauthor{finkel-etal-2008-efficient}~\shortcite{finkel-etal-2008-efficient} and \citeauthor{durrett-klein-2015-neural}~\shortcite{durrett-klein-2015-neural}.
% 后面再写

% 关于ACL-2019 Head-Driven Phrase Structure Grammar Parsing on Penn Treebank 这篇文章
% idea非常简单，基本就是一个联合解码。但是Section 2硬是和HPSG扯上关系，写得很复杂。
% 我们能用他们的代码跑出来差不多的结果吗？
% Table 1，division我理解就是baseline，分别解码的？为什么会这么差呢？比joint低了很多。
% 最基本的baseline：在他们的框架上，去单独训练依存或短语句法分析器，看看结果。看看到底是网络增大导致的性能提升，还是因为两个任务直接MTL，不兼容导致division结果很差。我们之前也试过，直接短语和依存MTL，如果不仔细做，很可能性能有大的下降。
% table 6你们在汉语短语句法上的实验结果，是否都是用了BERT的结果？我在文章中没有找到明确的说明。感觉像是用了BERT。
% 2）table 4在汉语依存句法上的结果，看着都挺正常的，好像没有用BERT，不过里面提到了两种standard data splitting，具体是什么？为什么用了短语的data split，性能就提高了很多？实验部分最开始只讲用了standard data split，突然就有两个呢？而且要和别人的结果对比，就要搞清楚并说清楚到底短语和依存的data split差异在哪。
% 头疼！要不要引用呢？看这个文章感觉脑子很乱。

Because of the inefficiency issue, there exist only a few previous works on CRF constituency parsing.
\citeauthor{finkel-etal-2008-efficient}~\shortcite{finkel-etal-2008-efficient} propose the first non-neural feature-rich CRF constituency parser.
\citeauthor{durrett-klein-2015-neural}~\shortcite{durrett-klein-2015-neural} extend the work of \citeauthor{finkel-etal-2008-efficient}~\shortcite{finkel-etal-2008-efficient} and use a feedforward neural network with non-linear activation for scoring anchored production rules.
Both works perform explicit inside-outside computations on CPUs and suffer from a severe inefficiency issue.

This work is built on the high-performance modern neural parser based on a BiLSTM encoder \cite{stern-etal-2017-minimal}, which first applies minus features \cite{cross-huang-2016-span} for span scoring to graph-based constituency parsing.
% and employ dynamic programming decoding for finding the 1-best tree.
Several recent works follow \citeauthor{stern-etal-2017-minimal}~\shortcite{stern-etal-2017-minimal}.
\citeauthor{gaddy-etal-2018-whats}~\shortcite{gaddy-etal-2018-whats} try to analyze what and how much context is implicitly encoded by BiLSTMs.
\citeauthor{kitaev-klein-2018-constituency}~\shortcite{kitaev-klein-2018-constituency} replace two-layer BiLSTM with self-attention layers and find considerable improvement via separated content and position attending.
In contrast, this work shows that the parser of \citeauthor{stern-etal-2017-minimal}~\shortcite{stern-etal-2017-minimal} outperforms \citeauthor{kitaev-klein-2018-constituency}~\shortcite{kitaev-klein-2018-constituency} by properly configuring  BiLSTMs such as the dropout strategy (see Table~\ref{table:dev}). Please also kindly notice that \citeauthor{kitaev-klein-2018-constituency}~\shortcite{kitaev-klein-2018-constituency} use very large word embeddings.

Batchification is straightforward and well-solved for sequence labeling tasks, as shown in the implementation of NCRF++\footnote{\url{https://github.com/jiesutd/NCRFpp}}.
However, very few works turned sight to tree-structures.
In a slightly earlier work, we for the first time propose to batchify tree-structured inside and Viterbi (Eisner) computation for GPU acceleration for the dependency parsing \cite{zhang-etal-2020-efficient}.
This work is an extension to the constituency parsing with different inside and Viterbi (CKY) algorithms.

As an independent and concurrent work to ours, Torch-Struct\footnote{\url{https://github.com/harvardnlp/pytorch-struct}}, kindly brought up by a reviewer, has also implemented batchified TreeCRF algorithms for constituency parsing \cite{alex-2020-torchstruct}.
%As the most closely related works, we both proposed to use the batchification techniques and deploy the auto-differentiation in place of outside for speeding up.
%However, there are some key differences between Torch-Struct and ours.
However, Torch-Struct aims to provide general-purpose basic implementations for structure prediction algorithms.
In contrast, we work on sophisticated parsing models, and aim to advance the state-of-the-art CRF constituency parsing in both accuracy and efficiency.

Meanwhile, there is a recent trend of extremely simplifying the constituency parsing task without explicit structural consideration or the use of CKY decoding.
%a growing body of research focuses on simplifying the parsing procedure.
%Consequently, many novel paradigms of constituency parsing have been proposed.
% \citeauthor{choe-charniak-2016-parsing}~\shortcite{choe-charniak-2016-parsing} 语言模型。
\citeauthor{gomez-rodriguez-vilares-2018-constituent}~\shortcite{gomez-rodriguez-vilares-2018-constituent} propose a sequence labeling approach for constituency parsing by designing a complex tag encoding tree information for each input word.
% tag包含两部分：数字表达分叉所在的层数；label表示分叉对应组块的标签
 %therefore the time complexity is reduced to $O(n)$.
\citeauthor{vilares-etal-2019-better}~\shortcite{vilares-etal-2019-better} further enhance the sequence labeling approach via several augmentation strategies such as multi-task learning and policy gradients.
\citeauthor{shen-etal-2018-straight}~\shortcite{shen-etal-2018-straight} propose to predict a scalar distance in the gold-standard parse tree for each neighboring word pairs and employ bottom-up greedy search to find an optimal tree.
However, all the above works lag behind the mainstream approaches by a large margin in terms of parsing performance.

\section{Conclusions}
\label{section:conclusions}

In this work, we propose a fast and accurate neural CRF constituency parser. We show that the inside and CKY algorithms can be effectively batchified to accommodate direct large tensor computation on GPU, leading to dramatic efficiency improvement.
The back-propagation procedure is equally efficient and erases the need for the outside algorithm for gradient computation.
Experiments on three English and Chinese benchmark datasets lead to several promising findings.
First, the simple two-stage bracketing-then-labeling approach is more efficient than one-stage parsing without hurting performance.
Second, our new scoring architecture achieves higher performance than the previous method based on minus features.
Third, the dropout strategy we introduce can improve parsing performance by a large margin.
Finally, our proposed parser achieves new state-of-the-art performances with a parsing speed of over 1,000 sentences per second.

% The efficiency weakness is well solved by utilizing a batchified inside algorithm.
% We empirically verify the equivalence of the outside algorithm and back-propagation.
% This naturally enables us to do the MBR decoding by using marginal probabilities as the byproduct of computing gradients.
% We borrow the architecture from the widely used biaffine parser in two main aspects: 1) using biaffine as our scoring mechanism in place of the minus-feature; 2) adopt a two-stage bracketing-then-labeling parsing strategy, leading to fast and accurate parsing.
% Our results have outperformed the self-attentive model and achieve the state-of-the-art in the case of not using any external embeddings.

\section*{Acknowledgments}

The authors would like to thank the anonymous reviewers for their helpful comments.
This work was supported by National Natural Science Foundation of China (Grant No. 61525205, 61876116) and a Project Funded by the Priority Academic Program Development (PAPD) of Jiangsu Higher Education Institutions.
%% The file named.bst is a bibliography style file for BibTeX 0.99c
\bibliographystyle{named}
\bibliography{ijcai20}

\begin{thebibliography}{}

\bibitem[\protect\citeauthoryear{Akoury \bgroup \em et al.\egroup
  }{2019}]{akoury-etal-2019-syntactically}
Nader Akoury, Kalpesh Krishna, and Mohit Iyyer.
\newblock Syntactically supervised transformers for faster neural machine
  translation.
\newblock In {\em Proceedings of ACL}, pages 1269--1281, 2019.

\bibitem[\protect\citeauthoryear{Collins}{1997}]{collins-1997-three}
Michael Collins.
\newblock Three generative, lexicalised models for statistical parsing.
\newblock In {\em Proceedings of ACL}, pages 16--23, 1997.

\bibitem[\protect\citeauthoryear{Cross and Huang}{2016}]{cross-huang-2016-span}
James Cross and Liang Huang.
\newblock Span-based constituency parsing with a structure-label system and
  provably optimal dynamic oracles.
\newblock In {\em Proceedings of EMNLP}, pages 1--11, 2016.

\bibitem[\protect\citeauthoryear{Devlin \bgroup \em et al.\egroup
  }{2019}]{devlin-etal-2019-bert}
Jacob Devlin, Ming-Wei Chang, Kenton Lee, and Kristina Toutanova.
\newblock {BERT}: Pre-training of deep bidirectional transformers for language
  understanding.
\newblock In {\em Proceedings of NAACL}, pages 4171--4186, 2019.

\bibitem[\protect\citeauthoryear{Dozat and
  Manning}{2017}]{Timothy-d17-biaffine}
Timothy Dozat and Christopher~D. Manning.
\newblock Deep biaffine attention for neural dependency parsing.
\newblock In {\em Proceedings of ICLR}, 2017.

\bibitem[\protect\citeauthoryear{Durrett and
  Klein}{2015}]{durrett-klein-2015-neural}
Greg Durrett and Dan Klein.
\newblock Neural {CRF} parsing.
\newblock In {\em Proceedings of ACL-IJCNLP}, pages 302--312, 2015.

\bibitem[\protect\citeauthoryear{Eisner}{2016}]{eisner-2016-inside}
Jason Eisner.
\newblock Inside-outside and forward-backward algorithms are just backprop
  (tutorial paper).
\newblock In {\em Proceedings of WS}, pages 1--17, 2016.

\bibitem[\protect\citeauthoryear{Finkel \bgroup \em et al.\egroup
  }{2008}]{finkel-etal-2008-efficient}
Jenny~Rose Finkel, Alex Kleeman, and Christopher~D. Manning.
\newblock Efficient, feature-based, conditional random field parsing.
\newblock In {\em Proceedings of ACL}, pages 959--967, 2008.

\bibitem[\protect\citeauthoryear{Gaddy \bgroup \em et al.\egroup
  }{2018}]{gaddy-etal-2018-whats}
David Gaddy, Mitchell Stern, and Dan Klein.
\newblock What{'}s going on in neural constituency parsers? an analysis.
\newblock In {\em Proceedings of NAACL}, pages 999--1010, 2018.

\bibitem[\protect\citeauthoryear{G{\'o}mez-Rodr{\'\i}guez and
  Vilares}{2018}]{gomez-rodriguez-vilares-2018-constituent}
Carlos G{\'o}mez-Rodr{\'\i}guez and David Vilares.
\newblock Constituent parsing as sequence labeling.
\newblock In {\em Proceedings of EMNLP}, pages 1314--1324, 2018.

\bibitem[\protect\citeauthoryear{Jin \bgroup \em et al.\egroup
  }{2020}]{jin-etal-2020-relation}
Lifeng Jin, Linfeng Song, Yue Zhang, Kun Xu, Wei yun Ma, and Dong Yu.
\newblock Relation extraction exploiting full dependency forests.
\newblock In {\em Proceedings of AAAI}, 2020.

\bibitem[\protect\citeauthoryear{Kaplan \bgroup \em et al.\egroup
  }{2004}]{kaplan-etal-2004-speed}
Ron Kaplan, Stefan Riezler, Tracy~H. King, John~T. Maxwell~III, Alex Vasserman,
  and Richard Crouch.
\newblock Speed and accuracy in shallow and deep stochastic parsing.
\newblock In {\em Proceedings of HLT-NAACL}, pages 97--104, 2004.

\bibitem[\protect\citeauthoryear{Kitaev and
  Klein}{2018}]{kitaev-klein-2018-constituency}
Nikita Kitaev and Dan Klein.
\newblock Constituency parsing with a self-attentive encoder.
\newblock In {\em Proceedings of ACL}, pages 2676--2686, 2018.

\bibitem[\protect\citeauthoryear{Kitaev \bgroup \em et al.\egroup
  }{2019}]{kitaev-etal-2019-multilingual}
Nikita Kitaev, Steven Cao, and Dan Klein.
\newblock Multilingual constituency parsing with self-attention and
  pre-training.
\newblock In {\em Proceedings of ACL}, pages 3499--3505, 2019.

\bibitem[\protect\citeauthoryear{Lample \bgroup \em et al.\egroup
  }{2016}]{lample-etal-2016-neural}
Guillaume Lample, Miguel Ballesteros, Sandeep Subramanian, Kazuya Kawakami, and
  Chris Dyer.
\newblock Neural architectures for named entity recognition.
\newblock In {\em Proceedings of NAACL}, pages 2475--2485, 2016.

\bibitem[\protect\citeauthoryear{Le and Zuidema}{2014}]{le-zuidema-2014-inside}
Phong Le and Willem Zuidema.
\newblock The inside-outside recursive neural network model for dependency
  parsing.
\newblock In {\em Proceedings of EMNLP}, pages 729--739, 2014.

\bibitem[\protect\citeauthoryear{Li \bgroup \em et al.\egroup
  }{2019}]{li-etal-2019-attentive}
Ying Li, Zhenghua Li, Min Zhang, Rui Wang, Sheng Li, and Luo Si.
\newblock Self-attentive biaffine dependency parsing.
\newblock In {\em Proceedings of IJCAI}, pages 5067--5073, 2019.

\bibitem[\protect\citeauthoryear{Matsuzaki \bgroup \em et al.\egroup
  }{2005}]{matsuzaki-etal-2005-probabilistic}
Takuya Matsuzaki, Yusuke Miyao, and Jun{'}ichi Tsujii.
\newblock Probabilistic {CFG} with latent annotations.
\newblock In {\em Proceedings of ACL}, pages 75--82, 2005.

\bibitem[\protect\citeauthoryear{Peters \bgroup \em et al.\egroup
  }{2018}]{peters-etal-2018-deep}
Matthew Peters, Mark Neumann, Mohit Iyyer, Matt Gardner, Christopher Clark,
  Kenton Lee, and Luke Zettlemoyer.
\newblock Deep contextualized word representations.
\newblock In {\em Proceedings of NAACL}, pages 2227--2237, 2018.

\bibitem[\protect\citeauthoryear{Petrov and
  Klein}{2007}]{petrov-klein-2007-improved}
Slav Petrov and Dan Klein.
\newblock Improved inference for unlexicalized parsing.
\newblock In {\em Proceedings of NAACL}, pages 404--411, 2007.

\bibitem[\protect\citeauthoryear{Rush}{2020}]{alex-2020-torchstruct}
Alexander~M. Rush.
\newblock Torch-struct: Deep structured prediction library.
\newblock arXiv:2002.00876, 2020.

\bibitem[\protect\citeauthoryear{Sagae and
  Lavie}{2005}]{sagae-lavie-2005-classifier}
Kenji Sagae and Alon Lavie.
\newblock A classifier-based parser with linear run-time complexity.
\newblock In {\em Proceedings of IWPT}, pages 125--132, 2005.

\bibitem[\protect\citeauthoryear{Shen \bgroup \em et al.\egroup
  }{2018}]{shen-etal-2018-straight}
Yikang Shen, Zhouhan Lin, Athul~Paul Jacob, Alessandro Sordoni, Aaron
  Courville, and Yoshua Bengio.
\newblock Straight to the tree: Constituency parsing with neural syntactic
  distance.
\newblock In {\em Proceedings of ACL}, pages 1171--1180, 2018.

\bibitem[\protect\citeauthoryear{Smith and
  Smith}{2007}]{smith-smith-2007-probabilistic}
David~A. Smith and Noah~A. Smith.
\newblock Probabilistic models of nonprojective dependency trees.
\newblock In {\em Proceedings of EMNLP}, pages 132--140, 2007.

\bibitem[\protect\citeauthoryear{Stern \bgroup \em et al.\egroup
  }{2017}]{stern-etal-2017-minimal}
Mitchell Stern, Jacob Andreas, and Dan Klein.
\newblock A minimal span-based neural constituency parser.
\newblock In {\em Proceedings of ACL}, pages 818--827, 2017.

\bibitem[\protect\citeauthoryear{Taskar \bgroup \em et al.\egroup
  }{2004}]{taskar-etal-2004-max}
Ben Taskar, Dan Klein, Mike Collins, Daphne Koller, and Christopher Manning.
\newblock Max-margin parsing.
\newblock In {\em Proceedings of EMNLP}, pages 1--8, 2004.

\bibitem[\protect\citeauthoryear{Teng and Zhang}{2018}]{teng-zhang-2018-two}
Zhiyang Teng and Yue Zhang.
\newblock Two local models for neural constituent parsing.
\newblock In {\em Proceedings of COLING}, pages 119--132, 2018.

\bibitem[\protect\citeauthoryear{Vilares \bgroup \em et al.\egroup
  }{2019}]{vilares-etal-2019-better}
David Vilares, Mostafa Abdou, and Anders S{\o}gaard.
\newblock Better, faster, stronger sequence tagging constituent parsers.
\newblock In {\em Proceedings of NAACL}, pages 3372--3383, 2019.

\bibitem[\protect\citeauthoryear{Wang and Chang}{2016}]{wang-chang-2016-graph}
Wenhui Wang and Baobao Chang.
\newblock Graph-based dependency parsing with bidirectional {LSTM}.
\newblock In {\em Proceedings of ACL}, pages 2475--2485, 2016.

\bibitem[\protect\citeauthoryear{Wang \bgroup \em et al.\egroup
  }{2018}]{wang-etal-2018-tree}
Xinyi Wang, Hieu Pham, Pengcheng Yin, and Graham Neubig.
\newblock A tree-based decoder for neural machine translation.
\newblock In {\em Proceedings of EMNLP}, pages 4772--4777, 2018.

\bibitem[\protect\citeauthoryear{Zhang \bgroup \em et al.\egroup
  }{2020}]{zhang-etal-2020-efficient}
Yu~Zhang, Zhenghua Li, and Zhang Min.
\newblock Efficient second-order {TreeCRF} for neural dependency parsing.
\newblock In {\em Proceedings of ACL}, 2020.

\bibitem[\protect\citeauthoryear{Zhou and Zhao}{2019}]{zhou-zhao-2019-head}
Junru Zhou and Hai Zhao.
\newblock Head-driven phrase structure grammar parsing on {P}enn treebank.
\newblock In {\em Proceedings of the ACL}, pages 2396--2408, 2019.

\bibitem[\protect\citeauthoryear{Zhu \bgroup \em et al.\egroup
  }{2013}]{zhu-etal-2013-fast}
Muhua Zhu, Yue Zhang, Wenliang Chen, Min Zhang, and Jingbo Zhu.
\newblock Fast and accurate shift-reduce constituent parsing.
\newblock In {\em Proceedings of ACL}, pages 434--443, 2013.

\end{thebibliography}
\end{document}